\pdfoutput=1

\documentclass[11pt]{article}

\usepackage[final]{acl}
\usepackage{pifont}

\usepackage{times}
\usepackage{latexsym}

\usepackage[T1]{fontenc}

\usepackage[utf8]{inputenc}

\usepackage{microtype}
\usepackage{amsmath}
\usepackage{inconsolata}

\usepackage{graphicx}

\usepackage{arydshln}
\usepackage{caption} 
\usepackage{amsmath}
\usepackage{amssymb}
\usepackage{xcolor}
\usepackage{subfigure}
\usepackage{multirow}
\usepackage[normalem]{ulem}

\definecolor{skyblue}{HTML}{D4F6FF}
\definecolor{lightred}{HTML}{FFE3E3}
\definecolor{lightbluegreen}{HTML}{D1F8EF}
\definecolor{lightpurple}{HTML}{EBEAFF}

\title{RankCoT: Refining Knowledge for Retrieval-Augmented Generation through Ranking Chain-of-Thoughts}

\author{Mingyan Wu$^{1}$, Zhenghao Liu$^{1}$\thanks{ \ \ indicates corresponding author.}, Yukun Yan$^{2}$\footnotemark[1],\\ \textbf{Xinze Li$^{1}$, Shi Yu$^{2}$, Zheni Zeng$^{2}$, Yu Gu$^{1}$, Ge Yu$^{1}$} \\ 
$^1$Department of Computer Science and Technology, Northeastern University, China \\
$^2$Department of Computer Science and Technology, Institute for AI, Tsinghua University, China \\
Beijing National Research Center for Information Science and Technology, China \\
}

\begin{document}
\maketitle
\begin{abstract}
Retrieval-Augmented Generation (RAG) enhances the performance of Large Language Models (LLMs) by incorporating external knowledge. However, LLMs still encounter challenges in effectively utilizing the knowledge from retrieved documents, often being misled by irrelevant or noisy information. To address this issue, we introduce RankCoT, a knowledge refinement method that incorporates reranking signals in generating CoT-based summarization for knowledge refinement based on given query and all retrieval documents. During training, RankCoT prompts the LLM to generate Chain-of-Thought (CoT) candidates based on the query and individual documents. It then fine-tunes the LLM to directly reproduce the best CoT from these candidate outputs based on all retrieved documents, which requires LLM to filter out irrelevant documents during generating CoT-style summarization. Additionally, RankCoT incorporates a self-reflection mechanism that further refines the CoT outputs, resulting in higher-quality training data. Our experiments demonstrate the effectiveness of RankCoT, showing its superior performance over other knowledge refinement models. Further analysis reveals that RankCoT can provide shorter but effective refinement results, enabling the generator to produce more accurate answers. All code and data are available at \url{https://github.com/NEUIR/RankCoT}.
\end{abstract}

\section{Introduction}
Retrieval-Augmented Generation (RAG)~\cite{NEURIPS2020_6b493230, guu2020retrieval, ram-etal-2023-context, shi-etal-2024-replug} empowers Large Language Models (LLMs) to access external knowledge, providing up-to-date information during the generation process. RAG models have demonstrated their effectiveness in mitigating the hallucination problem commonly encountered by LLMs~\cite{shuster2021retrieval}, enhancing the performance of LLMs, such as GPT-4~\cite{achiam2023gpt} and LLaMA~\cite{touvron2023llama}, in different NLP tasks.

\begin{figure}[t] 
\centering
    \includegraphics[width=0.5\textwidth]{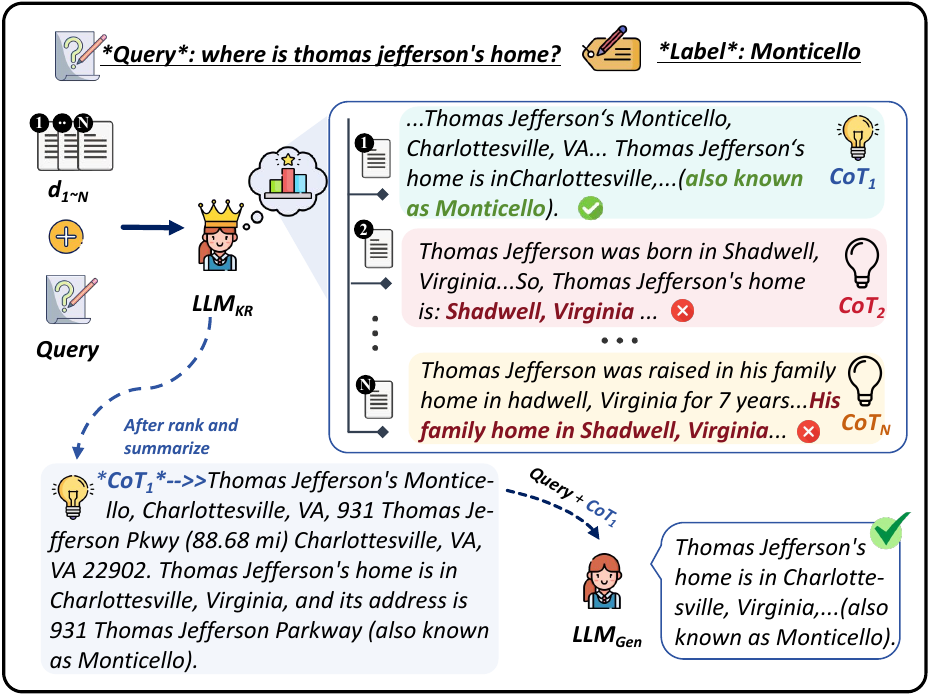}
    \caption{Illustration of RankCoT. We present how knowledge refinement can be achieved by incorporating reranking into CoT-based summarization.} \label{fig:RankCoT_intro}
\end{figure}
RAG models typically use dense retrieval methods~\cite{karpukhin2020dense, DBLP:conf/iclr/XiongXLTLBAO21} to retrieve query-relevant documents from external knowledge bases. These documents, along with the query, are then fed as the input context into LLMs~\cite{ram-etal-2023-context}. Thriving on their in-context learning capabilities~\cite{brown2020language,dong2022survey}, LLMs are able to extract relevant semantics from retrieved documents and generate appropriate responses to address the given query. However, the potential knowledge conflict between external knowledge and parameterized memory still poses a challenge for LLMs in generating precise responses~\cite{chen2024benchmarking, asai2024reliable}.

Many RAG models focus on building modular RAG pipelines to enhance retrieval performance~\cite{gao2024modular,asai2024selfrag}. These models primarily aim to refine the retrieved knowledge by assessing the relevance of each document to the query and subsequently filtering out irrelevant ones~\cite{yan2024corrective, asai2024selfrag}. However, the reranking model still requires feeding the remaining documents into LLMs, which means that query-unrelated content within a relevant document may still mislead the generators~\cite{xu2024recomp}. Some models address this problem by prompting LLMs to summarize query-relevant knowledge from the retrieved documents, thereby reducing the influence of irrelevant information~\cite{vig2021exploring, yu-etal-2024-chain, xu2024recomp}. This summarization approach often incorporates information from unrelated documents as part of the summaries, resulting in the introduction of noise. Both the reranking and summarization modules have advantages for knowledge refinement. However, in existing RAG systems, these modules are typically modeled separately by prompting the same LLMs.


This paper presents RankCoT, a knowledge refinement method that combines the strengths of both ranking and summarization to effectively enhance the process for retrieval result refinement. As shown in Figure~\ref{fig:RankCoT_intro}, we feed both the query and all retrieved documents into the RankCoT model, which incorporates reranking signals in generating CoT-based summarization as knowledge refinements, thereby aiding LLMs in generating more accurate responses for answering the given query. During training RankCoT, we independently feed the query and retrieved document to the LLM, asking it to generate several Chain-of-Thought (CoT) responses to answer the question, which can be considered as summarization results. We then design the self-refinement model to prompt LLMs to answer the question according to these sampled CoTs, helping to refine the CoT results for more effective training. If the refined CoT contains the ground truth answer, it is considered a positive refinement result, while those that do not contain the ground truth answer are considered negative refinements.

Our experiments demonstrate that RankCoT outperforms all baseline models, achieving over a 2\% improvement. Notably, RankCoT proves effective across LLMs of various scales. It generates shorter knowledge refinement results compared to both reranking and summarization methods, while enhancing the response accuracy of generator. Further analysis reveals that RankCoT successfully incorporates ground truth answers into the knowledge refinement results, while also including more query-relevant content. Additionally, RankCoT effectively extracts crucial semantics from the retrieved documents and alleviates the conflict between retrieved contents and internal knowledge.

\section{Related Work}
Retrieval-Augmented Generation (RAG) aims to enhance Large Language Models (LLMs) by enabling them to access external knowledge bases, providing up-to-date information during the generation process~\cite{shi-etal-2024-replug, ram-etal-2023-context}. This approach has demonstrated promising results across various NLP tasks, including open-domain question answering~\cite{izacard2022few}, code generation~\cite{zhou2023docprompting}, and dialogue~\cite{shuster2022blenderbot}. In these RAG models, retrieved documents are typically used as context to assist LLMs in generating more accurate responses~\cite{ram-etal-2023-context}. However, the conflict between the external knowledge and the parametric memory of LLMs often undermines the effectiveness of current RAG systems~\cite{asai2024reliable, xie2024adaptive, chen2024benchmarking}.

To mitigate the potentially negative impact of retrieved knowledge, existing models focus on refining the external knowledge through various modules designed to help LLMs generate more precise responses. Earlier works concentrate on reranking the retrieved documents~\cite{yu2023augmentation,shi-etal-2024-replug,yu2024rankrag}, while others employ query-focused summarization techniques~\cite{vig2021exploring,xu2023lmgqs} to reduce noise. However, reranking models often overlook noise within individual passages, and summarization models may fail to account for query-document relevance, sometimes incorporating misleading content in the summarization results. Chain-of-Note~\cite{yu-etal-2024-chain} attempts to instruct LLMs to generate query-related notes when answering a given query. This model incorporates the knowledge refinement process into the reasoning stage~\cite{wei2022chain} and heavily relies on the capabilities of LLMs, which may limit its applicability in RAG systems~\cite{gao2024modular}.

Modular RAG systems~\cite{gao2024modular,xu2024activerag} focus on refining external knowledge through different modules implemented by LLMs, which have become a key trend in the RAG area.
For instance, Self-RAG~\cite{asai2024selfrag} uses different tags for adaptive retrieval~\cite{jiang2023active} and self-reflection to refine knowledge. Some approaches also focus on reformulating queries to identify more useful documents for answering questions~\cite{yan2024corrective,trivedi2023interleaving}. \citet{yan2024corrective} introduce a retrieval evaluator that acts as a judge to trigger query reformulation, search, and knowledge refinement actions to supply more accurate evidence for generation.

To further improve the performance of modular RAG systems, these models focus on fine-tuning various components of the RAG framework. Some efforts aim to align the information needs between the retriever and the generator by optimizing the retrievers based on feedback from the generation models~\cite{yu2023augmentation, shi-etal-2024-replug, izacard2021distilling}. \citet{lin2024radit} adapt LLMs within the RAG setting by constructing instruction-tuning data for Supervised Fine-Tuning (SFT), enabling the models to better leverage the retrieved documents. Additionally, \citet{li2024rag} use Direct Preference Optimization (DPO)~\cite{rafailov2024direct} to jointly optimize the modules in a RAG system, aligning their data preferences.

\section{Methodology}
\begin{figure*}[t] 
\centering
    \includegraphics[width=1.0\textwidth]{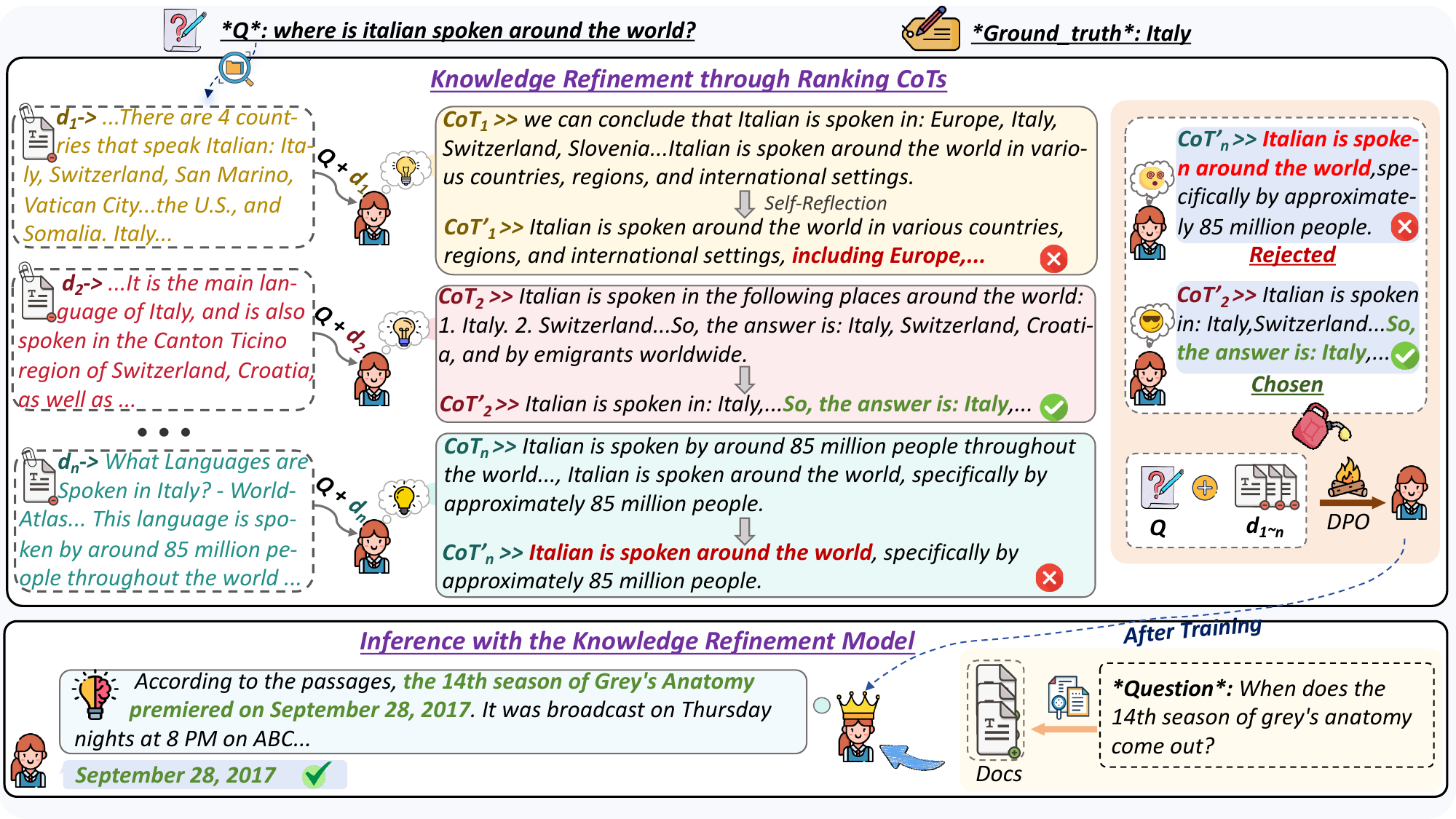}
    \caption{Illustration of RankCoT.} \label{fig:RankCoT}
\end{figure*}

As illustrated in Figure~\ref{fig:RankCoT}, this section introduces the RankCoT method. First, we introduce the preliminary of knowledge refinement in Retrieval-Augmented Generation (RAG) systems (Sec.~\ref{sec:preliminary}). And then we describe how to optimize LLMs to produce more effective chain-of-thoughts for knowledge refinement (Sec.~\ref{sec:rankcot}).

\subsection{Preliminary of Knowledge Refinement in Retrieval-Augmented Generation Systems}\label{sec:preliminary}
Given a query $q$ and a set of retrieved documents $D = \{d_1, d_2, \ldots, d_n\}$, vanilla RAG system~\cite{ram-etal-2023-context} uses retrieved documents as context and leverages the in-context learning method to help the generation model $\mathcal{M}_\text{Gen}$ produce the answer $y_\text{Gen}$:
\begin{equation}
    (q, D) \rightsquigarrow \mathcal{M}_\text{Gen} \rightsquigarrow y_\text{Gen}.
\end{equation}
Instead of directly feeding retrieved documents to the generation model ($\mathcal{M}_\text{Gen}$), some RAG models~\cite{gao2024modular, asai2024selfrag} design various modules to refine retrieved documents $D$:
\begin{equation}\label{eq:refinement}
    (q,D) \rightsquigarrow \mathcal{M}_\text{KR} \rightsquigarrow y_\text{KR}.
\end{equation}
The refined knowledge $y_\text{KR}$ is then passed to the generation model to mitigate the negative impact of retrieval noise:
\begin{equation}
    (q, y_\text{KR}) \rightsquigarrow \mathcal{M}_\text{Gen} \rightsquigarrow y_\text{Gen}.
\end{equation}
In the rest of this subsection, we will introduce different methods for implementing the knowledge refinement model $\mathcal{M}_\text{KR}$ in Eq.~\ref{eq:refinement}, including reranking, summarization, and RankCoT.

\textbf{Reranking.} Following previous work~\cite{asai2024selfrag}, we prompt LLMs to evaluate the relevance of the $i$-th retrieved document $d_i$ with respect to the query $q$, and output a binary label $y_\text{Rerank}^i$ for filtering out noisy documents:
\begin{equation}
\label{eq:reranking}
    y_\text{Rerank}^i = \text{LLM} (\text{Instruct}_\text{Rerank}, q, d_i),
\end{equation}
where $\text{Instruct}_\text{Rerank}$ prompts the LLM to assess the relevance between $q$ and $d_i$. The prediction label $y_\text{Rerank}^i$ can be ``YES'' or ``NO'', indicating whether the $i$-th document $d_i$ is relevant or irrelevant to the query $q$. We then retain the documents predicted as ``YES'' and construct the filtered document set $\{d_1, \dots, d_k\}$. The knowledge refinement result $y_\text{KR}$ is then represented as:
\begin{equation}
    y_\text{KR} = d_1 \oplus \dots \oplus d_k,
\end{equation}
where $\oplus$ is the concatenation operation.

\textbf{Summarization.} Another approach for knowledge refinement is summarization, which aims to extract query-related content from the retrieved documents~\cite{vig2021exploring}. The knowledge refinement result can be obtained as:
\begin{equation}
    y_\text{KR} = \text{LLM} (\text{Instruct}_\text{Sum}, q, D),
\end{equation}
where $\text{Instruct}_\text{Sum}$ is the instruction prompting the LLM to generate a summary. Unlike reranking, summarization directly generates the refined knowledge, avoiding the need to feed raw documents to the generation model $\mathcal{M}_\text{Gen}$.

\textbf{RankCoT.} RankCoT further incorporates a Chain-of-Thought (CoT)~\cite{wei2022chain} into the knowledge refinement process:
\begin{equation}
    y_\text{KR} = \text{LLM} (\text{Instruct}_\text{CoT}, q, D),
\end{equation}
where $\text{Instruct}_\text{CoT}$ is the instruction that prompts the LLM to generate CoT. RankCoT incorporates the chain-of-thought reasoning as the knowledge refinement result $y_\text{KR}$ to extract relevant knowledge from retrieved documents $D$, thereby assisting RAG models in answering the query.  Unlike summarization, RankCoT integrates the reranking mechanism during the chain-of-thought generation (Sec.~\ref{sec:rankcot}) to mitigate the influence of noisy documents~\cite{liu2024lost, xie2024adaptive}.

\subsection{Knowledge Refinement through Ranking Chain-of-Thoughts}\label{sec:rankcot}
To generate tailored knowledge refinement results $y_\text{KR}$ for RAG modeling, RankCoT optimizes the LLM ($\mathcal{M}$) by incorporating reranking into the CoT generation process. Furthermore, we introduce a self-reflection method to further refine the CoT results, mitigating the risk of overfitting to undesired CoT patterns during training RankCoT.

\subsubsection{Reranking Modeling in CoT Generation}
To learn the query-document relevance, we feed each document $d_i$ into the LLM ($\mathcal{M}$) and sample one CoT output $y_\text{CoT} (d_i)$:
\begin{equation}
\label{eq:CoTrank}
 y_\text{CoT} (d_i) \sim \mathcal{M} (q, d_i).
\end{equation}
Next, we gather all generated CoT results from each document in the retrieved document set $D$ to form the candidate CoT set $Y_\text{CoT}$:
\begin{equation}
 Y_\text{CoT} =  \{y_\text{CoT} (d_1), \dots, y_\text{CoT} (d_n)\}.
\end{equation}
We treat the CoT result $y_\text{CoT} \in Y_\text{CoT}$ that contains the ground truth answer as the positive $y_\text{CoT}^+$, while the result that does not contain the ground truth answer is regarded as the negative $y_\text{CoT}^-$.

Finally, we can optimize the LLM ($\mathcal{M}$) to assign higher generation probabilities to positive knowledge refinement results $y_\text{CoT}^+$ than the negative ones $y_\text{CoT}^-$. The training process is implemented with the Direct Preference Optimization (DPO) method~\cite{rafailov2024direct}:
\begin{equation}
\begin{aligned}
\label{eq:dpo}
& \mathcal{L} = 
- \mathbb{E}_{(q, y_\text{CoT}^+,y_\text{CoT}^-) \sim \mathcal{T}} \Big[ \log \sigma \Big(  \beta \log \\
&\frac{\mathcal{M}(y_\text{CoT}^+ \mid  q, D)}{\mathcal{M}^\text{Ref}(y_\text{CoT}^+ \mid  q, D)} - 
\beta \log \frac{\mathcal{M}(y_\text{CoT}^- \mid q, D)}{\mathcal{M}^\text{Ref}(y_\text{CoT}^- \mid q, D)} \Big) \Big],
\end{aligned}
\end{equation}
where $\beta$ is a hyperparameter and $\sigma$ is the Sigmoid function. $\mathcal{M}^\text{Ref}$ is the reference model, which remains frozen during training.

In DPO training, we input all documents $D$ into the model $\mathcal{M}$ and aim to assign a higher probability to the positive knowledge refinement result $y_\text{CoT}^+$, which is individually generated from one of the retrieved documents. This guides the model $\mathcal{M}$ to rerank the retrieved documents $D$ when generating chain-of-thought as the refinement.

\subsubsection{CoT Refinement through Self-Reflection}
While these generated CoTs help the RAG model generate more accurate answers, the generated CoT results of LLMs may contain undesired patterns, such as ``According to the document'' and ``the reasoning process is''. These training patterns can mislead the LLM ($\mathcal{M}$) to overfit these CoT results during training~\cite{gudibande2023false}. To address this problem, RankCoT proposes a self-reflection method to refine the CoT results $Y_\text{CoT}$.

Specifically, we first sample the CoT outputs $\Tilde{y}_\text{CoT} (d_i)$ by feeding the given query $q$ and each document $d_i$ to the LLM: 
\begin{equation}
 \Tilde{y}_\text{CoT} (d_i) \sim \mathcal{M} (\text{Instruct}_\text{CoT}, q, d_i),
\end{equation}
where $\text{Instruct}_\text{CoT}$ is used to prompt the LLM to generate a chain-of-thought. Then the CoT result $\Tilde{y}_\text{CoT} (d_i)$ is refined as $y_\text{CoT} (d_i)$ by using the same LLM ($\mathcal{M}$):
\begin{equation}
 y_\text{CoT} (d_i) = \mathcal{M} (\text{Instruct}_\text{Ref}, q, \Tilde{y}_\text{CoT} (d_i)),
\end{equation}
where the instruction $\text{Instruct}_\text{Ref}$ prompts the LLM ($\mathcal{M}$) to answer the given query $q$ based on the initial CoT result $\Tilde{y}_\text{CoT} (d_i)$. Such a self-reflection mechanism helps to extract more query-related contents from the initial CoT $\Tilde{y}_\text{CoT} (d_i)$, producing higher-quality data to optimize LLMs.
Finally, we collect the refined CoT results to form $Y_\text{CoT} (d_i) = \{y_\text{CoT} (d_1), \dots, y_\text{CoT} (d_n)\}$, which is used to train the LLM through DPO (Eq.~\ref{eq:dpo}).

\section{Experimental Methodology}
In this section, we describe the datasets, baselines, evaluation metrics, and implementation details in our experiments. More experimental details are shown in Appendix~\ref{appendix:aed}.

\textbf{Datasets.}
In our experiments, we follow previous work~\cite{lin2024radit} and utilize the instruction tuning datasets to train and evaluate RAG models. For all datasets and baselines, we use BGE-large~\cite{chen2024bge} to retrieve documents from the MS MARCO V2.1 document collection~\cite{bajaj2016ms}. We select six datasets for evaluation, including NQ~\cite{kwiatkowski2019natural}, HotpotQA~\cite{yang2018hotpotqa}, Trivia QA~\cite{joshi2017triviaqa}, PopQA~\cite{mallen-etal-2023-trust}, ASQA~\cite{stelmakh-etal-2022-asqa}, and MARCO QA~\cite{bajaj2016ms}, which require models to retrieve factual knowledge or conduct more complex reasoning to help answer the given query. All data statistics are shown in Table~\ref{data statistics simple}.

\textbf{Baselines.}
In our experiments, we compare RankCoT with the Vanilla RAG (No Refinement) model and three knowledge refinement models, including Rerank, Summary, and CoT, which are described in Sec.~\ref{sec:preliminary}. For the vanilla RAG model, we follow previous work~\cite{ram-etal-2023-context} and feed 5 retrieved documents as context to answer the question. For the Rerank model, we prompt the LLM to evaluate the relevance between the query and retrieved documents~\cite{asai2024selfrag, li2024rag}. If the document is relevant to the question, it outputs ``YES'' and retains the document, otherwise, it outputs ``NO'' and discards the document. The Summary model and CoT model prompt the LLM to extract query-related knowledge from retrieved documents using summarization and Chain-of-Thought~\cite{wei2022chain} formats to conclude query-related knowledge from retrieved documents.

\begin{table}[t]
  \centering
  \begin{tabular}{l|rl}
    \hline
    \textbf{Split} &\textbf{\#Query} &\textbf{Tasks}\\
    \hline
    \multirow{2}{*}{Train} &{15,444} &{Open-Domain QA}\\
    &{8,527} &{Reasoning}\\
    \hdashline
    \multirow{2}{*}{Dev} &{1,752} &{Open-Domain QA}\\
    &{912} &{Reasoning}\\
    \hdashline
    Test &{20,294} &{QA}\\
    \hline
  \end{tabular}
  \caption{\label{data statistics simple}Data Statistics.}
\end{table}

\textbf{Evaluation Metrics.} Following~\citet{xu2024unsupervised}, we utilize Rouge-L as evaluation metric for MARCO QA task. Following~\citet{gao-etal-2023-enabling}, we utilize String-EM as evaluation metric for ASQA. For other tasks, we use the Accuracy metric for evaluation.

\textbf{Implementation Details.}
We implement our RankCoT model using Llama3-8b-Instruct~\cite{touvron2023llama} as the backbone model. To construct a training dataset for DPO training, we ask Llama3-8b-Instruct to generate CoTs using 10 retrieved documents independently and use the same model to refine CoTs. During training, we feed 5 relevant documents as external knowledge and ask the LLM to reproduce the refined CoT results. We use LoRA~\cite{hulora} method to fine-tune Llama3-8B-Instruct, with $\beta$ set to 0.1 and the learning rate set to 2e-5.

For the RAG model, we concatenate the generated CoT with the query to let Llama3-8B-Instruct generate the final answer. In addition, we also use LLMs of different scales, such as MiniCPM3-4B~\cite{hu2024minicpm} and Qwen2.5-14B-Instruct~\cite{yang2024qwen2}, to build the RAG model and evaluate the generalization ability of RankCoT.

\begin{table*}
  \centering
  \begin{tabular}{lccccccc}
    \hline
    \multirow{2}{*}{\textbf{Method}} & \textbf{NQ} & \textbf{HotpotQA} & \textbf{TriviaQA} & \textbf{PopQA} &\textbf{ASQA} &  \textbf{MARCO} &\textbf{Avg.} \\
    &{(acc)} &{(acc)} &{(acc)} &{(acc)} &{(str-em)} &{(rouge)}\\
    \hline
    \multicolumn{8}{l}{\textit{Llama3-8B-Instruct}} \\
    \hdashline
    No Refinement    &{45.68} & {29.43} & {82.85} &{35.60} & {38.79}  & {20.73} &{42.18}\\
    Rerank & {46.18} &{30.30} &{83.51}  &{36.20} &  {39.92} & {20.72} &{42.81}\\
    Summary     &{44.27} & {28.05} & {82.09} &{33.67} & {37.81} &\textbf{22.67} &{41.32}\\
    CoT   &{45.33} & {26.36} & {81.45} &{34.13}  &{40.25}   & {19.52} &{41.17}\\
    RankCoT  & \textbf{47.41} & \textbf{32.21} & \textbf{85.18} &\textbf{41.17} & \textbf{41.02}  &{20.84} &\textbf{44.64}\\
    \hline
    \multicolumn{8}{l}{\textit{MiniCPM3-4B}} \\
    \hdashline
    No Refinement    &{42.51} &{24.93} &{80.91} &{32.53} &{24.31}  & {13.55} &{36.46}\\
    RankCoT  & \textbf{48.78} & \textbf{33.13} & \textbf{85.20} &\textbf{36.87}  &\textbf{35.85} & \textbf{24.59} &\textbf{44.07}\\
    \hline
    \multicolumn{8}{l}{\textit{Qwen2.5-14B-Instruct}} \\
    \hdashline
    No Refinement    &{47.66} & {29.70} & {79.49} &{36.97} &\textbf{44.73} & {18.50} &{42.84}\\
    RankCoT  & \textbf{49.98} & \textbf{33.91} & \textbf{86.68} &\textbf{44.45}  & {41.94} & \textbf{24.62} &\textbf{46.93}\\
    \hline
  \end{tabular}
  \caption{\label{main result}
    Overall Performance of RAG System with Different Knowledge Refinement Models. We use Llama3-8B-Instruct as the backbone model for different knowledge refinement models and apply RankCoT to the RAG system, which is implemented with Llama3-8B-Instruct, MiniCPM3-4B, and Qwen2.5-14B-Instruct. 
  }
\end{table*}
\section{Evaluation Result}
In this section, we first evaluate the performance of various RAG methods, followed by ablation studies to examine the impact of self reflection module and different training strategies. We then investigate the characteristics of RankCoT by analyzing the knowledge utilization capabilities of RAG models using different knowledge refinement models. We also examine the effectiveness of the refined knowledge generated by RankCoT through answering consistency in Appendix~\ref{appendix:selfconsistency}. The case study is conducted in Appendix~\ref{appendix:casestudy}.

\subsection{Overall Performance}
This section presents the knowledge refinement performance of different models, as shown in Table~\ref{main result}. Additional baseline comparison results are shown in Appendix~\ref{appendix: more_baseline}.

The evaluation results reveal that these three knowledge refinement models, Rerank, Summary, and CoT, show distinct performance. Specifically, Rerank exhibits a slight improvement, while both Summary and CoT lead to a decrease in RAG performance. This highlights the challenge of effectively refining knowledge for RAG modeling. In contrast, RankCoT demonstrates a 2.5\% improvement over vanilla RAG model, indicating its effectiveness in providing more meaningful refinements that help LLMs better answer questions. Furthermore, RankCoT outperforms the Rerank model with a 1.8\% improvement and avoids the need to feed raw passages into the LLM twice for knowledge refinement and question answering.

Then we present the performance of RankCoT by applying it to different RAG systems implemented with LLMs of various scales. The results indicate that RankCoT maintains its effectiveness across different RAG configurations, yielding a 7.6\% improvement over the MiniCPM3-4B-based RAG model and a 4.1\% improvement over the RAG model implemented with Qwen2.5-14B-Instruct. These findings demonstrate that RankCoT has strong generalization ability in knowledge refinement, enabling LLMs of different scales to effectively leverage external knowledge.

\begin{table}[t]
  \centering
  \resizebox{\linewidth}{!}{
  \begin{tabular}{lcccc}
    \hline
    \multirow{2}{*}{\textbf{Method}}   & \textbf{NQ} & \textbf{HotpotQA} & \textbf{TriviaQA} &\multirow{2}{*}{\textbf{Avg.}}\\
    &{(acc)} &{(acc)} &{(acc)}\\
    \hline
    Vanilla RAG &{45.68} & {29.43} & {82.85} &{52.65}\\
    \hline
    \multicolumn{5}{l}{\textit{Inference w/ Finetuned QA Model}} \\
    \hdashline
    Rerank  & {45.96} & {29.88} & {83.04} &{52.96}\\
    Summary   &{38.46} &{25.29} &{76.13} &{45.63}\\
    CoT   &{41.63} & {33.23} &{83.24} &{52.70}\\
    \hline
    \multicolumn{5}{l}{\textit{SFT Training}} \\
    \hdashline
    Rerank  &{47.02} &{31.21} &{84.16} &{54.13}\\
    Summary   & {43.36} & {28.39} & {82.26} &{51.34}\\
    CoT   &{43.43} & {27.23} & {82.80} &{51.15}\\
    \hline
    \multicolumn{5}{l}{\textit{DPO Training}} \\
    \hdashline
    Rerank  & {46.80} & {30.70} & {84.70} &{54.07}\\
    Summary  &{46.63} & {30.61} & {83.95} &{53.73}\\
    RankCoT  &\textbf{47.41} & \textbf{32.21} & \textbf{85.18} &\textbf{54.97}\\
    w/o Reflect &{46.70} &{30.59} &{83.82} &{53.70}\\
    \hline
  \end{tabular}
  }
  \caption{\label{ablation study result}
    Ablation Study. Both SFT and DPO methods optimize knowledge refinement models using self-reflection labels. RankCoT w/o Reflect refers to that the RankCoT model is optimized using unrefined CoT.
  }
\end{table}
\begin{table*}[ht]
\centering
\small
\resizebox{\linewidth}{!}{
\begin{tabular}{l|ccc|ccc|ccc}
\hline
\multirow{2}{*}{\textbf{Method}} & \multicolumn{3}{c|}{\textbf{Has-Answer}} & \multicolumn{3}{c|}{\textbf{Miss-Answer}} & \multicolumn{3}{c}{\textbf{Internal Knowledge}}\\ 
& \textbf{NQ}  & \textbf{HotpotQA}  & \textbf{TriviaQA} & \textbf{NQ}  & \textbf{HotpotQA} & \textbf{TriviaQA} & \textbf{NQ}  & \textbf{HotpotQA} & \textbf{TriviaQA}\\ 
\hline
LLM w/o RAG &{46.41} 	&{42.99}	&{80.48}  &{6.51}	&{14.73}  &{46.04}  &100.0 &100.0  &100.0  \\
Vanilla RAG &{63.45} 	&{57.69}	&{88.76}  &{2.41}	&{8.89}  &{19.70}  &{77.30} &{67.89} &{90.80}  \\ \hdashline
Rerank &{64.09} 	&{57.09}	&{88.74}  &{3.14}	&{10.87}  &{27.41}  &{78.20} &{68.28} &{91.24}  \\
Summary &{61.40} 	&{53.20}	&{87.20}  &{3.02}	&{10.19}  &{25.27}  &{76.89} &{63.23} &{89.93}  \\
CoT &{62.90} 	&{51.76}	&{87.53}  &{2.53}	&{7.69}  &{22.91}  &{77.19} &{59.77} &{89.38}  \\\hdashline
RankCoT &\textbf{65.44} 	&\textbf{58.92}	&\textbf{90.49}  &\textbf{4.10}	&\textbf{13.03}  &\textbf{30.84}  &\textbf{80.42} &\textbf{70.81} &\textbf{92.61}  \\
\hline
\end{tabular}}
\caption{\label{table:scenarios} RAG Performance by Using Different Knowledge Refinement Models. We conduct three testing scenarios to evaluate the knowledge usage of RAG systems, including Has-Answer, Miss-Answer and Internal Knowledge.}
\end{table*}
\subsection{Ablation Study}
We conduct ablation studies to evaluate the effectiveness of various training strategies.

As shown in Table~\ref{ablation study result}, we first conduct the approach from prior work~\cite{lin2024radit}, where a fine-tuned QA model is used for knowledge refinement. We then train several knowledge refinement models--Rerank, Summary, and CoT--using the self-reflection mechanism introduced by RankCoT. Specifically, the self-reflection mechanism involves feeding the knowledge refinement results into LLMs to generate self-reflection results based on these inputs. In this setup, SFT methods select self-reflection results containing ground truth answers to train knowledge refinement models, while DPO methods select both positive and negative responses--those that contain ground truth answers and those that do not--for training.

After fine-tuning with QA supervision, the LLM is able to generate knowledge refinement results using different prompts. However, the evaluation results illustrate that these QA models present limited effectiveness compared to vanilla RAG model. We then train the knowledge refinement models using training signals refined by the LLM itself. Among the SFT-based models, Rerank achieves the best performance, illustrating that the reranking signals can be easily learned by LLMs through SFT. Using the DPO training method, both Summary and RankCoT show significant improvements over these knowledge refinement models using the SFT strategy. Furthermore, RankCoT outperforms all knowledge refinement models, demonstrating its effectiveness in producing effective knowledge refinement results to help LLMs generate more accurate answers. By replacing self-refined CoTs with the raw CoT outcomes during DPO training, RankCoT achieves a 1.3\% decline, showing the effectiveness of our self-reflection mechanism.

\subsection{Knowledge Usage Performance of Different Refinement Models}
In this experiment, we evaluate the ability of RankCoT to assist RAG models in leveraging external knowledge to generate final answers. We compare RankCoT with three knowledge refinement models: Rerank, Summary, and CoT.

As shown in Table~\ref{table:scenarios}, we conduct three testing scenarios to assess the effectiveness of the different knowledge refinement models: Has-Answer, Miss-Answer, and Internal Knowledge. The Has-Answer scenario involves cases where the retrieved documents contain the correct (golden) answer. This scenario evaluates whether the knowledge refinement model can effectively extract key information from these documents to aid the LLM in answering the question. The Miss-Answer scenario, on the other hand, deals with cases where the retrieved documents do not include the golden answer. This scenario further tests the ability of the knowledge refinement models to minimize the impact of retrieved noise. Finally, the Internal Knowledge scenario examines the ability of different knowledge refinement models to handle conflicts between internal and external knowledge.

As shown in the evaluation results, RankCoT outperforms all knowledge refinement models across all datasets in the Has-Answer scenario. This demonstrates the effectiveness of RankCoT in incorporating more query-relevant information from retrieved documents, thereby enhancing the accuracy of the RAG system in this scenario. In the Miss-Answer scenario, the performance of vanilla RAG models significantly drops compared to LLMs w/o RAG, indicating that query-irrelevant documents mislead LLMs into producing incorrect answers. However, the use of different knowledge refinement models mitigates this performance decline. Among these models, RankCoT exhibits the most substantial improvements, demonstrating its effectiveness in filtering out noisy information from retrieval and reducing the misleading information of irrelevant documents. In the internal knowledge scenario, knowledge refinement models--Rerank, Summary, and CoT--perform comparably or even worse than vanilla RAG model, illustrating that existing methods are less effective in addressing the knowledge conflict issue. In contrast, RankCoT outperforms these knowledge refinement models, demonstrating its ability to provide more tailored knowledge refinement results. The RankCoT-produced knowledge refinement results effectively alleviate knowledge conflicts, aiding LLMs in better utilizing both internal and external knowledge.


\begin{figure}[t]
    \centering
    \subfigure[Similarity between Query and Refined Knowledge.] { \label{fig:textsimquery} 
    \includegraphics[width=0.48\linewidth]{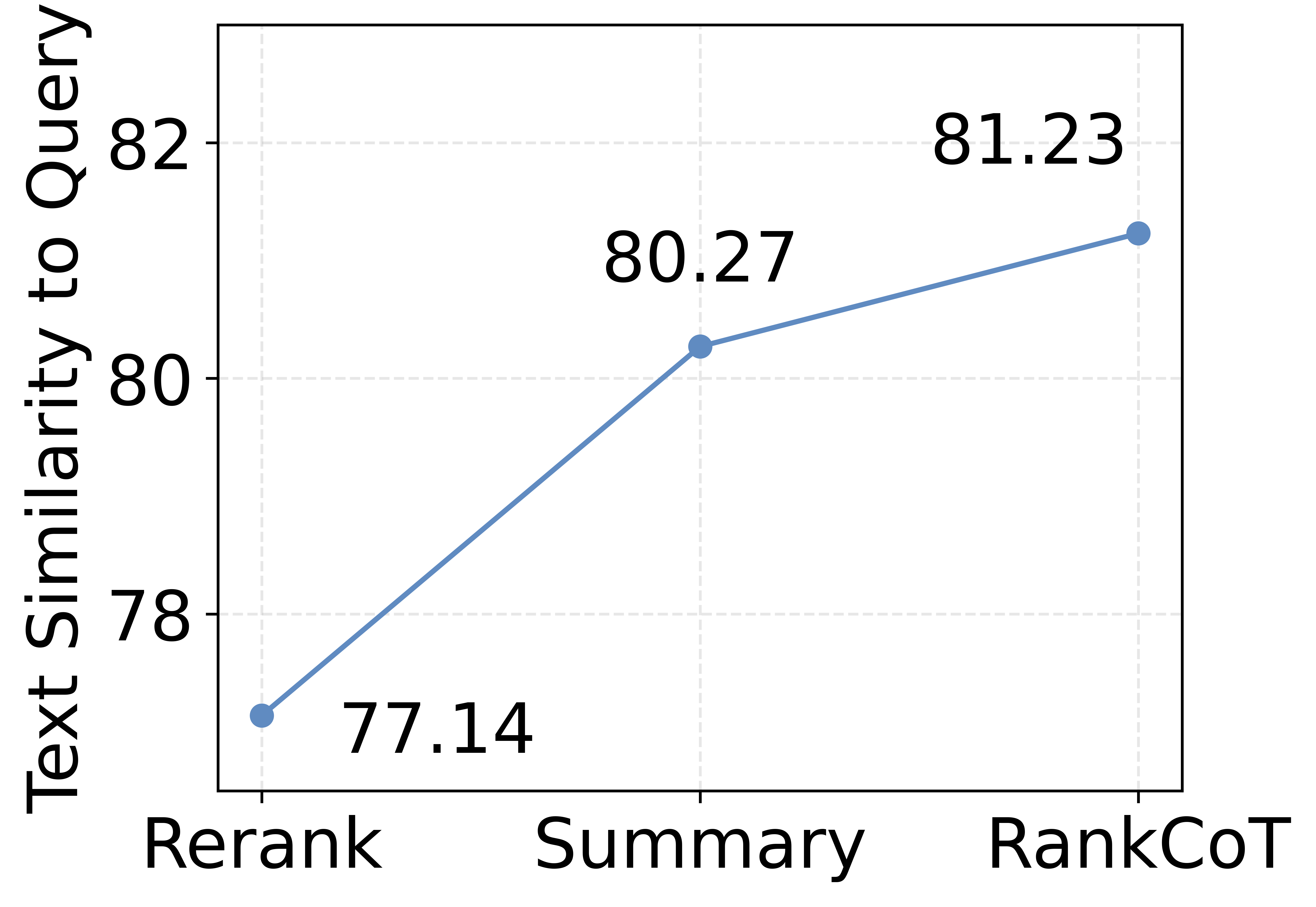}}
    \subfigure[Hit Rate of Ground Truth Answers.] { \label{fig:groundtruthin} 
    \includegraphics[width=0.48\linewidth]{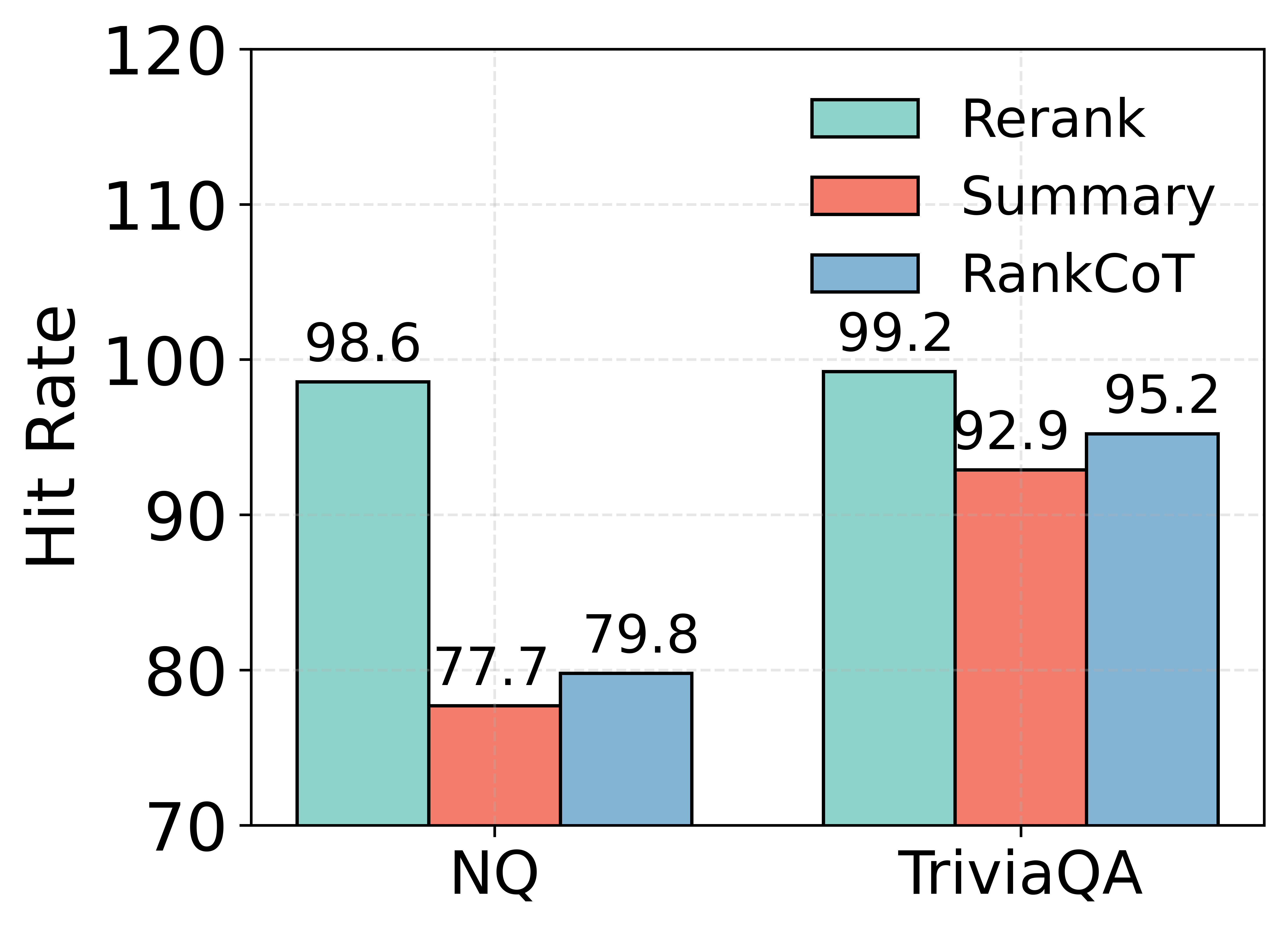}}
    \caption{Quality of Refined Knowledge Generated by Different Models. In Figure~\ref{fig:textsimquery}, we first estimate the text similarity between the query and the knowledge refinement results using the BGE model~\cite{chen2024bge}. Then, we calculate the hit rate of these knowledge refinement results in Figure~\ref{fig:groundtruthin}, which evaluates whether the ground truth answers are included in the knowledge refinement results.}
    \label{fig:refined knowledge quality}
\end{figure}
\begin{figure}[t]
    \centering
    \subfigure[Average Length.] { \label{fig:inputlengthlog} 
    \includegraphics[width=0.48\linewidth]{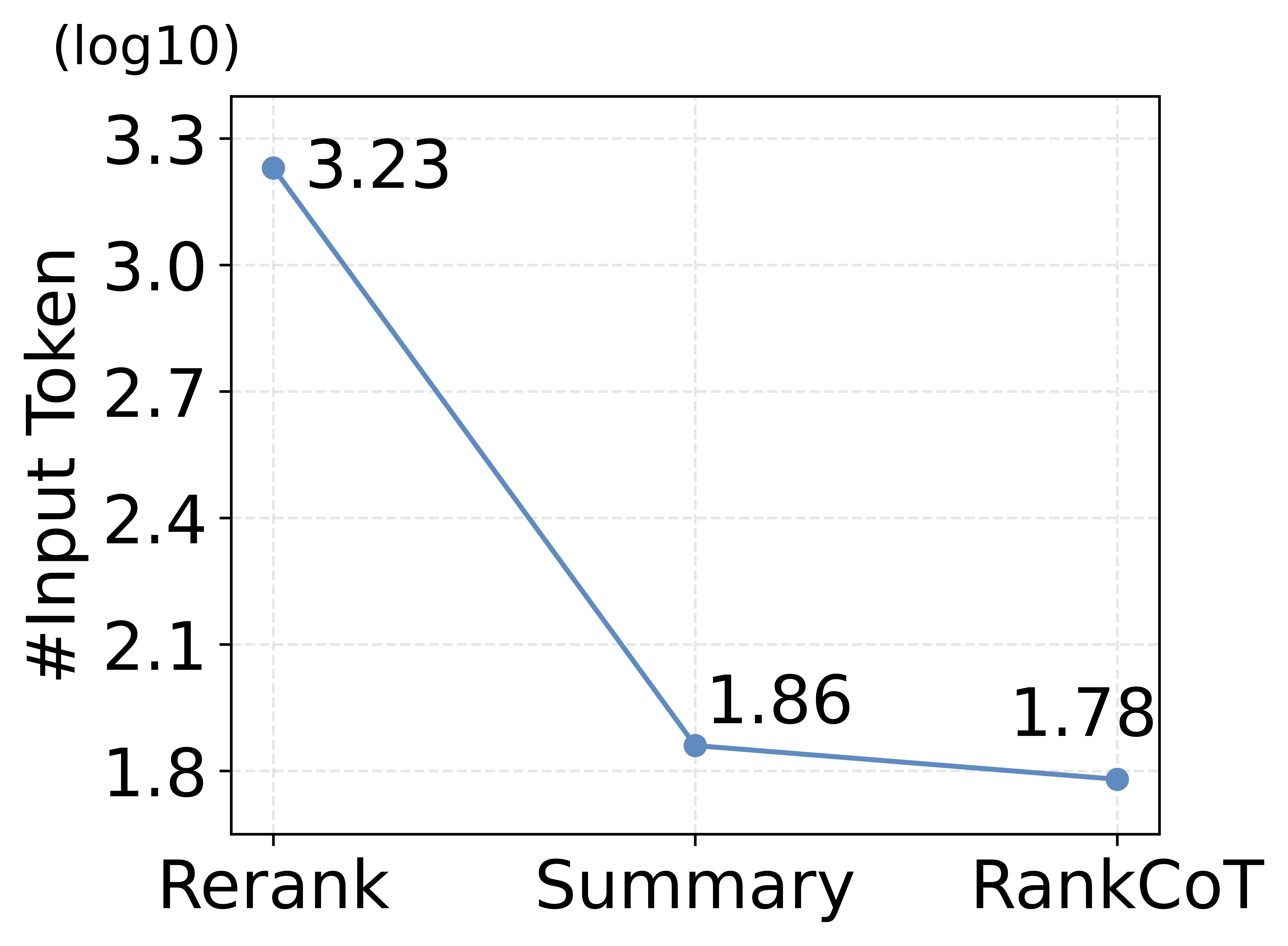}}
    \subfigure[Length Change Ratio.] { \label{fig:inputlengthratio} 
    \includegraphics[width=0.48\linewidth]{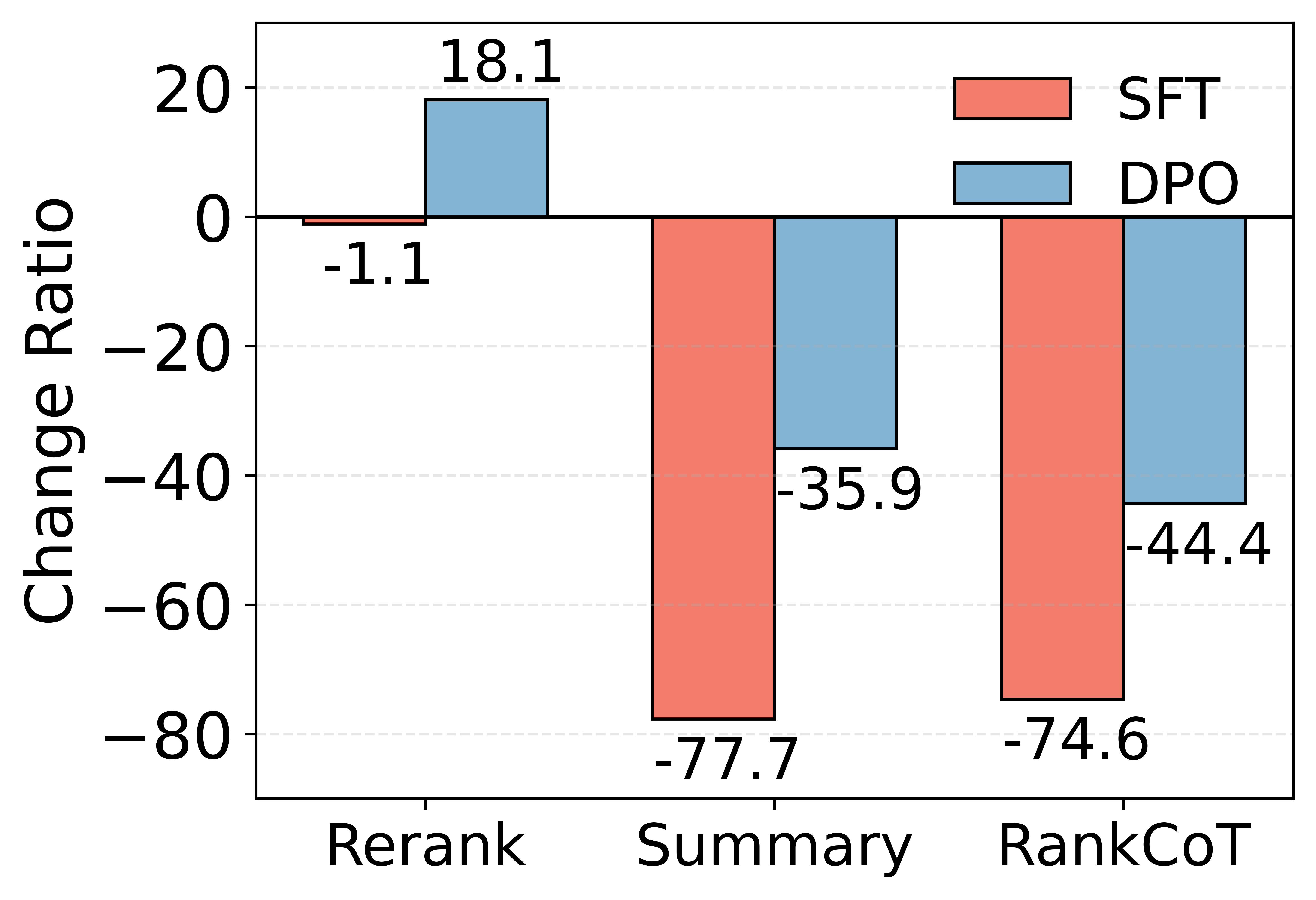}}
    \caption{The Length of Knowledge Refinement Results Produced by Different Models. We first present the average length of the refinement results in Figure~\ref{fig:inputlengthlog}. Then, the length change ratio relative to vanilla LLMs is illustrated in Figure~\ref{fig:inputlengthratio}.
    }
    \label{fig:inputlength}
\end{figure}

\subsection{Characteristics of the Refined Knowledge Produced by RankCoT}\label{sec:characteristics}
This experiment further explores the characteristics of RankCoT-produced knowledge refinement results by estimating both the quality and length of the refined knowledge. We conduct the experiment on the NQ and TriviaQA datasets, specifically in the Has-answer test scenario, where the retrieved documents contain the ground truth answers.

\textbf{Refinement Quality.} As shown in Figure~\ref{fig:refined knowledge quality}, we evaluate the quality of the knowledge refinement results based on query relevance and the hit rate of the golden answer. Specifically, the similarity scores between the query and the refined knowledge generated by three different knowledge refinement models are presented in Figure~\ref{fig:textsimquery}. RankCoT achieves the highest similarity score with the query, demonstrating its effectiveness in retaining more query-related information from the retrieved documents. Furthermore, we show the hit rate of the ground truth answer in Figure~\ref{fig:groundtruthin}. As indicated by the evaluation results, Rerank achieves the highest hit rates, while Summary performs the worst. This outcome likely stems from the fact that the Rerank model only selects the most relevant document, whereas the Summary model must extract key information, inevitably discarding some of the relevant contents that contain the ground truth answers. Although RankCoT is also a summarization-style knowledge refinement model, it achieves higher hit rates, showing that RankCoT can capture more ground truth answers in its refinement results.

\textbf{Length of Knowledge Refinement.} Subsequently, we present the results of knowledge refinement lengths for different models in Figure~\ref{fig:inputlength}. As shown in Figure~\ref{fig:inputlengthlog}, the summarization-style knowledge refinement methods, Summary and RankCoT, significantly reduce the length of the refined knowledge compared to the Rerank model. Notably, RankCoT achieves the shortest refinement length, demonstrating its effectiveness in minimizing the consumption of prompt inputs for LLMs~\cite{mu2023learning}. Additionally, we investigate the length change ratio across different training methods in Figure~\ref{fig:inputlengthratio}. As shown in the results, these SFT methods generally result in shorter knowledge refinement outputs, illustrating that SFT encourages the summarization-style knowledge refinement model to overfit training signals~\cite{li2024rag}. In contrast, the DPO training method helps these knowledge refinement models produce longer results, facilitating more flexible responses that incorporate more crucial knowledge.

\section{Conclusion}
This paper proposes RankCoT, a knowledge refinement method that leverages the strengths of both ranking and summarization to effectively refine the knowledge from retrieval results, thereby aiding LLMs in generating more accurate responses. Our experimental studies show that RankCoT can effectively refine external knowledge and balance the utilization of internal and external knowledge. In-depth analysis reveals that the CoT generated by our method has a high similarity to the query and a high ground truth answer hit rate. 
\section*{Limitations}
Although RankCoT demonstrates its effectiveness in refining retrieved knowledge for RAG systems, the quality of the refinement is still constrained by the capabilities of LLMs. Specifically, RankCoT is optimized using the DPO method, which relies on LLMs to generate meaningful chosen and rejected pairs during optimization. Therefore, the generation of meaningful preference pairs for optimization still heavily depends on the performance of the LLMs. Additionally, RankCoT (Llama3-8B-Instruct) can be applied to different RAG systems that implemented with LLMs of varying scales and show its effectiveness. The improvements may be diminished when larger-scale LLMs are used as the generation model of RAG systems, due to the stronger knowledge refinement capabilities of LLMs of a larger scale. This further highlights the importance of aligning the parameter scale of the LLM used to build the RankCoT model with that of the generation model in RAG systems.


\bibliography{references_norm}
\clearpage

\appendix
\begin{table}
  \centering
  \resizebox{\linewidth}{!}{
  \begin{tabular}{lcccc}
    \hline
    \multirow{2}{*}{\textbf{Method}} & \textbf{NQ} & \textbf{HotpotQA} & \textbf{TriviaQA} &\multirow{2}{*}{\textbf{Avg.}} \\
    &{(acc)} &{(acc)} &{(acc)}\\
    \hline
    Vanilla RAG    &{45.68} & {29.43} & {82.85} &{52.65}\\
    Self-RAG~\citeyearpar{asai2024selfrag} &{39.90} &{24.59} &{78.35} &{47.61}\\
    Recomp~\citeyearpar{xu2024recomp} &{40.47} &{25.36} &{79.04} &{48.29} \\
    SEGENC~\citeyearpar{vig2021exploring} &{40.47} &{24.89} &{76.73} &{47.36} \\
    RankCoT  & \textbf{47.41} & \textbf{32.21} & \textbf{85.18}  &\textbf{54.93}\\
    \hline
  \end{tabular}}
  \caption{\label{table:morebaselines}Overall Performance of More Baselines.}
\end{table}
\begin{table*}[t]
\begin{center}
\small
\begin{tabular}{l|l|l|l|l|c}
\hline
\textbf{Split}  &\textbf{Task}   &\textbf{Dataset} &\textbf{Metric}   &\textbf{Raw}  &\textbf{Filtered}\\ \cline{1-6}
\multirow{10}{*}{Training} & \multirow{6}{*}{Open-Domain QA}    
&  Commonsense QA~\citeyearpar{talmor2018commonsenseqa}   & Accuracy &4,000  &3,037\\
&& Math QA~\citeyearpar{amini2019mathqa}  & Accuracy &4,000 &3,509\\
&& Web Questions~\citeyearpar{berant2013semantic}   & Accuracy &3,578  &1,810\\
&& Wiki QA~\citeyearpar{yang2015wikiqa} &  Rouge-L &840 &840  \\
&& Yahoo! Answers QA &  Rouge-L &4,000 &4,000 \\
&& MARCO QA~\citeyearpar{bajaj2016ms}&  Rouge-L &4,000  &4,000\\ 
\cline{2-6}

&\multirow{4}{*}{Reasoning}
& Algebra QA with Rationales~\citeyearpar{ling2017program} & Accuracy &2,527 &2,222 \\
&&  Explanations for CommonsenseQ~\citeyearpar{aggarwal2021explanations}  & Accuracy &4,000  &3,202\\
&& Grade School Math 8K~\citeyearpar{cobbe2021training}  & Accuracy &4,000 &3,090 \\
&& StrategyQA~\citeyearpar{geva2021did} &Accuracy &1,860 &925 \\
\cline{1-6}

\multirow{6}{*}{Evaluation}  & \multirow{6}{*}{QA} &  
Natural Questions~\citeyearpar{kwiatkowski2019natural}  &Accuracy & 2,837 &-\\   
&&  HotpotQA~\citeyearpar{yang2018hotpotqa}   &Accuracy & 5,600  &-\\
& & TriviaQA~\citeyearpar{joshi2017triviaqa}    &Accuracy & 5,359 &-\\   
& &  PopQA~\citeyearpar{mallen-etal-2023-trust} &Accuracy & 3,000  &-\\
& &  ASQA~\citeyearpar{stelmakh-etal-2022-asqa} &STR-EM & 948  &-\\
& &  MARCO QA~\citeyearpar{bajaj2016ms} &Rouge-L & 3,000  &-\\
\cline{2-6}
\hline
\end{tabular}
\caption{Data Statistics.}
\label{table:dataset}
\end{center}
\end{table*}
\section{Appendix}

\subsection{License} \label{license}
This section summarizes the licenses of the datasets used in our experiments.

All of these datasets under their respective licenses and agreements allow for academic use: Natural Questions (CC-BY-SA-3.0 License); PopQA, Commonsense QA, Wiki QA, MARCO QA, StrategyQA, and Grade School Math 8K (MIT License); Web Questions and HotpotQA (CC-BY-4.0 License); TriviaQA, ASQA, Algebra QA with Rationales, and Math QA (Apache 2.0 License); Explanations for CommonsenseQ (CDLA-Sharing-1.0 License); Yahoo! Answers QA shows its terms of use at website\footnote{\url{https://tensorflow.google.cn/datasets/community_catalog/huggingface/yahoo_answers_qa}}.



\subsection{Additional Baseline Comparison Results}\label{appendix: more_baseline}
This section presents the comparison results between RankCoT and several baseline models. 

In this experiment, we compare RankCoT with four baselines: vanilla LLM, Self-RAG, Recomp, and SEGENC. Self-RAG~\citep{asai2024selfrag} optimizes Llama3-8B-Instruct to retrieve documents on demand and ranks them by reflecting the retrieved documents using reflection tokens. SEGENC~\cite{vig2021exploring} is a Query-Focused Summarization model, initialized from the BART model~\cite{lewis-etal-2020-bart}, which summarizes documents based on a given query. Recomp~\cite{xu2024recomp} proposes a method to compress retrieved documents, reducing the computational overhead of language models during inference.

As shown in Table~\ref{table:morebaselines}, Self-RAG, Recomp, and SEGENC all show performance degradation compared to vanilla RAG. This indicates that ranking, compressing, or summarizing documents inevitably lead to information loss, thereby reducing response accuracy. In contrast, RankCoT not only incorporates advantages of ranking and summarization, but also generates a CoT that preserves as much useful information as possible. This approach reduces the input length for the QA model while improving response accuracy.

\begin{figure}[t] 
\centering
    \includegraphics[width=0.45\textwidth]{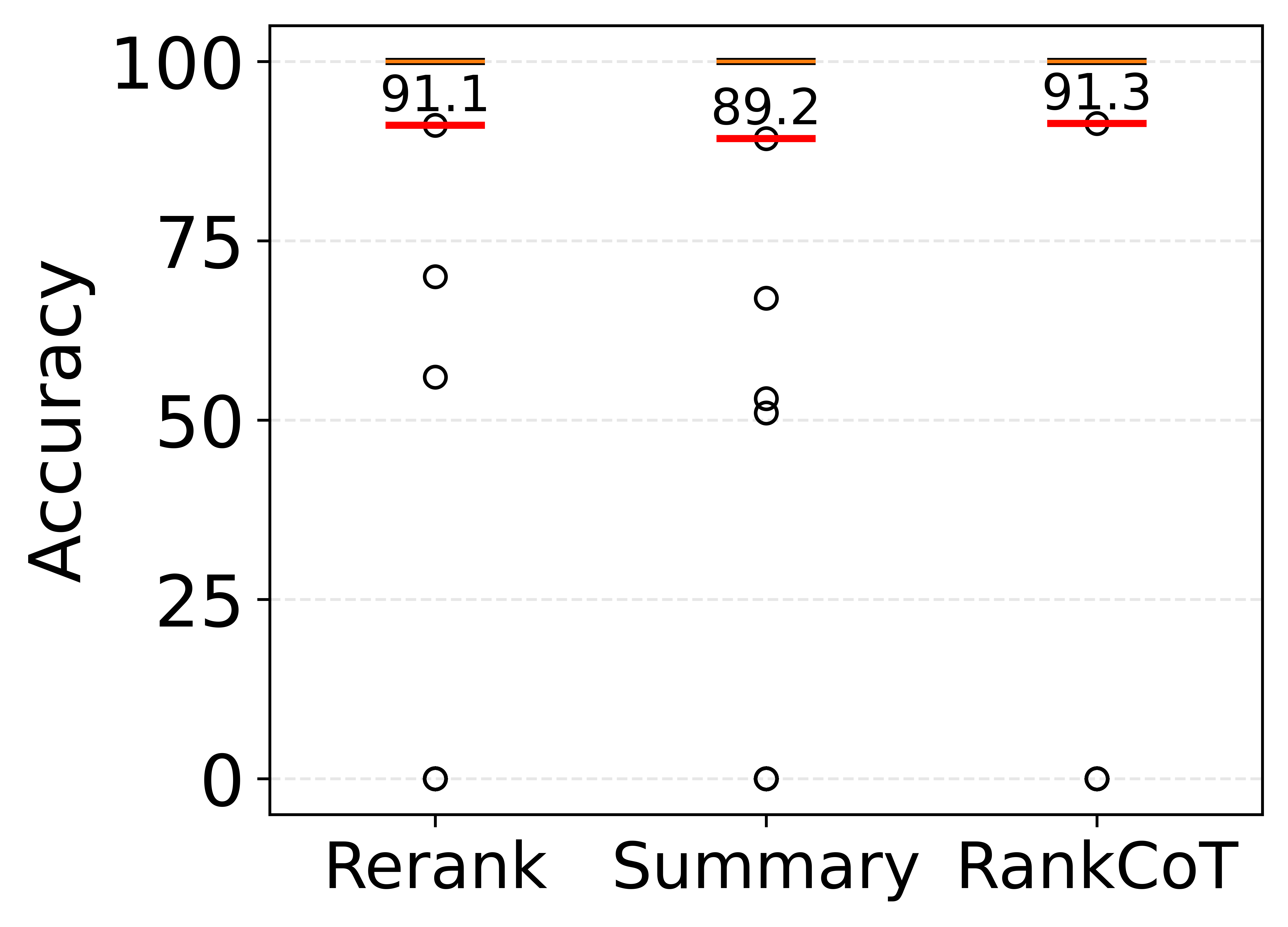}
    \caption{QA Consistency of the RAG Model Using Different Knowledge Refinement Models.} \label{fig:selfconsistency}
\end{figure}


\subsection{QA Consistency Using Different Knowledge Refinements}\label{appendix:selfconsistency}
This experiment evaluates the QA consistency based on different knowledge refinement results. Since the experiment is conducted in the internal knowledge scenario, all queries can be accurately answered by the RAG model without any external knowledge. 

As illustrated in Figure~\ref{fig:selfconsistency}, we use the TriviaQA dataset to conduct the experiment. For each query, we input both the query and the refinement results from different models, then sample responses from the RAG model 300 times. The ratio of correct answers is calculated and denoted as Accuracy, which serves to evaluate the QA consistency of the RAG model. A higher accuracy reflects that the knowledge refinement results help the RAG model consistently produce correct answers. 

As shown in the evaluation results, RankCoT demonstrates its effectiveness by achieving an average accuracy of approximately 91.3\%, outperforming both refinement baselines, Rerank and Summary. The accuracy of the Rerank and Summary methods shows significant variation, indicating that the knowledge refinements produced by both models still contain knowledge that causes the RAG model to lack consistency in its answers. In contrast, after applying RankCoT, the accuracy becomes concentrated at either 0 or 1, demonstrating that it better supports the RAG model in maintaining answer consistency.

\subsection{Additional Experimental Details}\label{appendix:aed} In this subsection, we first describe the process of constructing the training data and then show the prompt templates used in our experiments.

\textbf{Data Preprocessing for RankCoT.} The quantities of our training and evaluation data, along with the corresponding evaluation metrics, are presented in Table~\ref{table:dataset}. The ``Filtered'' column indicates the number of training samples used for DPO training.

During RankCoT training, we collect ten datasets, obtain 32,805 samples, and process them as described in Section~\ref{sec:rankcot}. Since the generated CoT can either be fully correct or incorrect, we cannot form preference data pairs from these generated CoT candidates. Thus, we filter out such samples and divide the remaining ones into training and validation datasets with a 9:1 ratio.

\begin{figure*}[t]
    \centering
    \subfigure[CoT Generation.] { \label{fig:CoTGenerationPrompt} 
    \includegraphics[width=0.8\textwidth]{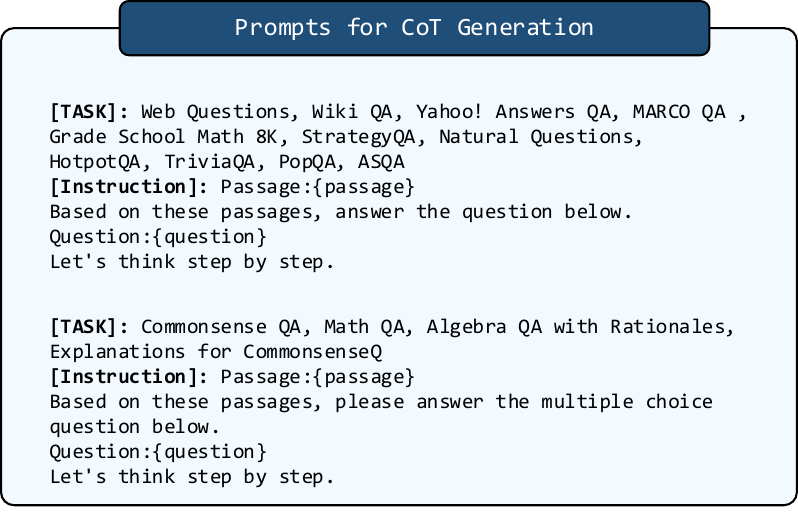}}
    \subfigure[CoT Refinement.] { \label{fig:CoTRefinementPrompt} 
    \includegraphics[width=0.8\textwidth]{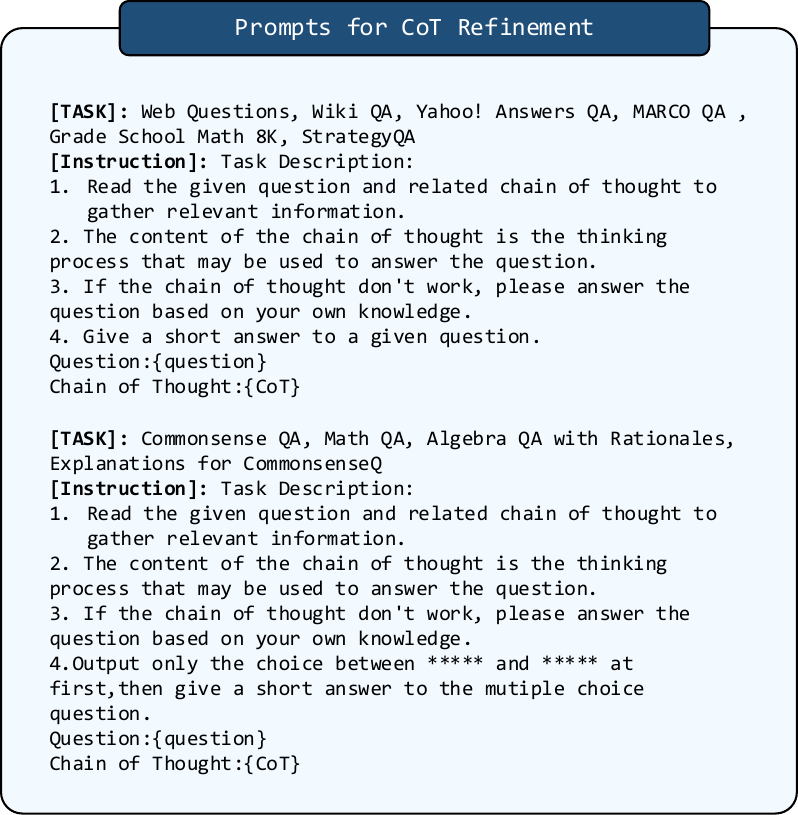}}
    \caption{Prompt Templates Used in RankCoT.
    }
    \label{fig:cotprompt}
\end{figure*}

\begin{figure*}[t] 
\centering
    \includegraphics[width=0.8\textwidth]{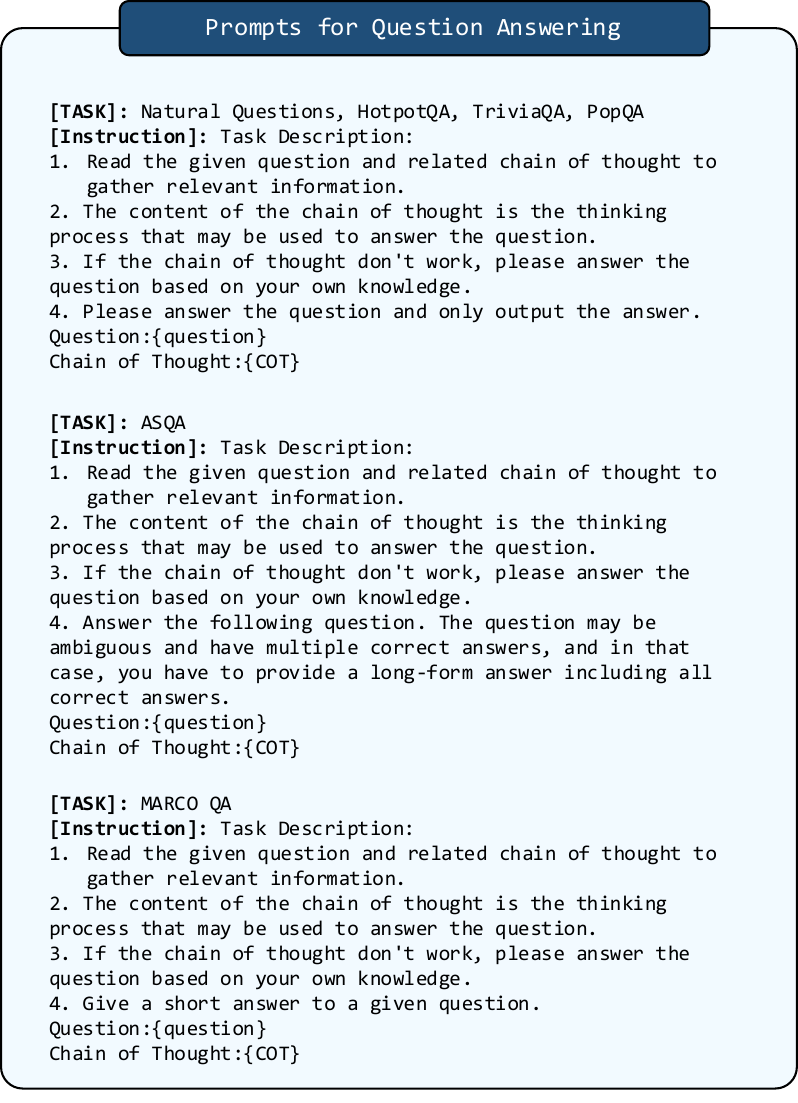}
    \caption{Prompt Templates Used for Question Answering.} \label{fig:QuestionAnsweringPrompt}
\end{figure*}

\begin{figure*}[t] 
\centering
    \includegraphics[width=0.8\textwidth]{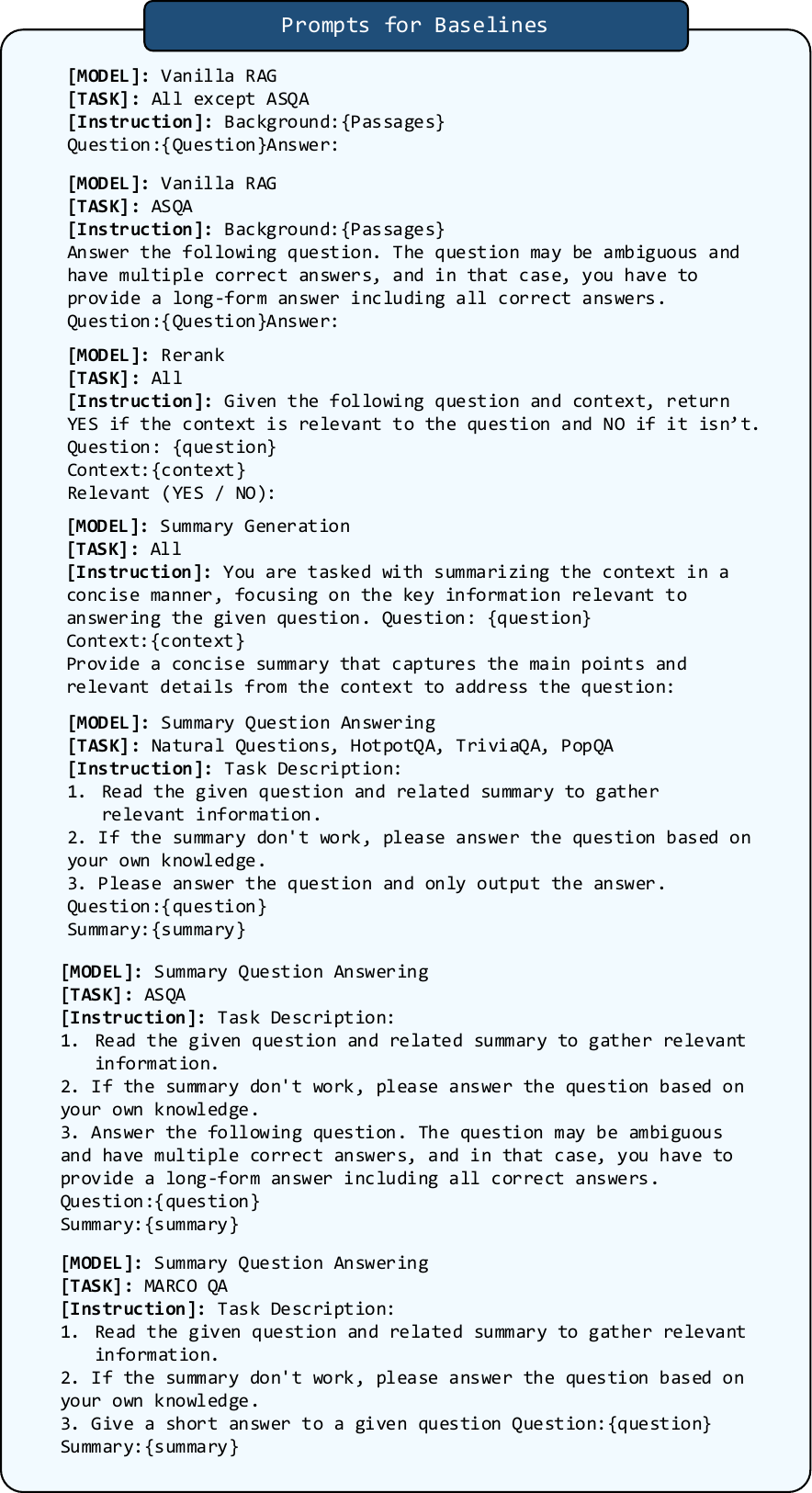}
    \caption{Prompt Templates Used for Implementing Baselines.} \label{fig:baselineprompt}
\end{figure*}

\textbf{Prompt Templates.} The prompts used for CoT generation ($\text{Instruct}_\text{CoT}$) are shown in Figure~\ref{fig:CoTGenerationPrompt}. The prompt used for question answering is illustrated in Figure~\ref{fig:QuestionAnsweringPrompt}. Additionally, the prompts used for CoT refinement ($\text{Instruct}_\text{Ref}$) are shown in Figure~\ref{fig:CoTRefinementPrompt}. Finally, Figure~\ref{fig:baselineprompt} presents the prompts used for implementing baselines.


\begin{table*}[ht]
  \centering
  \setlength{\fboxsep}{0.7pt}
  \resizebox{\linewidth}{!}{
  \begin{tabular}{p{1\textwidth}}
    \hline
    \textbf{Query:} What is australia's location in the world and region?    \; \textbf{Ground Truth: Oceania } \\
    \textbf{Documents:}\\

    (1) \ldots Australia is both a continent and a country. It is located to the south of Indonesia and Papua New Guinea, in the southeast part of the Indian Ocean. New Zealand lies to its east, and most of the \colorbox{skyblue}{Indian Ocean lies to its west}. \ldots Australia is the world’s \colorbox{lightred}{smallest } continent, though it is also the \colorbox{lightred}{sixth-largest} country in the world. The total land area of Australia is \colorbox{lightred}{7,682,300 sq. km}.\\

    (2) \ldots Australia is the smallest continent and the largest country in \colorbox{skyblue}{Oceania} located between the \colorbox{skyblue}{Indian Ocean and Pacific Ocean} in the \colorbox{skyblue}{Southern hemisphere}. \ldots Australia is separated from Papua New Guinea by the \colorbox{skyblue}{Coral Sea} and Torres Strait to the \colorbox{skyblue}{northeast}; from Indonesia by the \colorbox{skyblue}{Timor and Arafura} seas to the northwest; from Coral Sea Islands Territory by the Great Barrier Reef; from New Zealand by Tasman Sea to the southeast;\\

    (3) \ldots Australia is located \colorbox{skyblue}{south of the equator} and is bordered by the Indian Ocean and the Pacific Ocean.
    \ldots\\

    \hline
    \multicolumn{1}{c}{\textbf{Summary:}} \\
    \textbf{Summary:} Here is a concise summary: Australia is located in the \colorbox{lightred}{Southern Hemisphere}, between the Indian Ocean and the Pacific Ocean. It is the world's \colorbox{lightred}{smallest} continent and the \colorbox{lightred}{sixth-largest} country, covering an area of \colorbox{lightred}{7,682,300 square kilometers}. Australia is surrounded by the Indian Ocean to the west, the Timor, Arafura, and Coral Seas to the north, the Pacific Ocean to the east, and the Tasman Sea and Southern Ocean to the south. The country is made up of six states (Western Australia, South Australia, Victoria, Tasmania, New South Wales, and Queensland) and two territories (the Northern Territory and the Australian Capital Territory) \\
    \textbf{Answer:} Australia is located in the Southern Hemisphere, between the Indian Ocean and the Pacific Ocean. \textcolor{red}{(\ding{55})}\\
    \hline
    \multicolumn{1}{c}{\textbf{RankCoT:}} \\
    \textbf{Chain-of-Thought:} According to the passages, Australia is:
    
    1. An island continent located in the \colorbox{skyblue}{Southern Hemisphere}.
    
    2. Situated between the \colorbox{skyblue}{Indian Ocean and the Pacific Ocean}.
    
    3. Located \colorbox{skyblue}{south of the equator}.
    
    4. Surrounded by the \colorbox{skyblue}{Indian Ocean to the west}, the \colorbox{skyblue}{Timor, Arafura, and Coral Seas to the north}, the Pacific Ocean to the east, and the Tasman Sea and Southern Ocean to the south.
    
    5. An island continent \colorbox{skyblue}{in Oceania, which is a region} that includes Australia, New Zealand, and many Pacific island nations. 
    
    So, Australia's location in the world is in the Southern Hemisphere, \colorbox{lightpurple}{in the region of Oceania}.\\
    \textbf{Answer:} Australia is an island continent located in the Southern Hemisphere, in the \colorbox{lightpurple}{region of Oceania}. \textcolor{green}{(\ding{51})}\\
    \hline
    
  \end{tabular}
  }
  \caption{\label{case study}
    Case Study. We randomly sample one case from the NQ dataset to show the knowledge refinement result. Different colors are used to annotate the key information from the retrieved knowledge retained by different knowledge refinement models: \colorbox{lightred}{Pink} for Summary, \colorbox{skyblue}{Blue} for RankCoT. And we also highlight the key points that can help answer the query in \colorbox{lightpurple}{Purple}.
  }
\end{table*}

\subsection{Case Study}\label{appendix:casestudy}
In Table~\ref{case study}, we present a case study to illustrate the effectiveness of the RankCoT model. For the given query, it asks about ``Australia's location in the world and region''. And the retrieved documents contain both related and unrelated information about the geographical location of Australia.

The summarization method captures some geographical information about Australia from the retrieved documents, such as ``being in the Southern Hemisphere and located between the Indian and Pacific Oceans''. However, these descriptions offer only a broad geographical scope rather than directly answering the query about the regional location of Australia. Consequently, the LLM is misled by the ambiguous information in the summarized documents and generates inaccurate answers. Moreover, the summarization contains some irrelevant information, such as ``7,682,300 square kilometers'' and ``smallest continent and the sixth-largest country''. 

In contrast, RankCoT refines the retrieved documents in a question-answering based summarization manner by generating a Chain-of-Thought (CoT). These CoT results are constructed by sequentially integrating information from different retrieved documents, while ranking and prioritizing the most query-relevant knowledge. Rather than directly summarizing keypoint information from retrieved documents, RankCoT identifies crucial geographical attributes in each document and organizes them in a structured reasoning result. At the start of the CoT, broad geographical attributes, such as ``Southern Hemisphere'' and ``between the Indian and Pacific Oceans'' are strengthened, as they appear consistently across documents. More specific regional information, such as ``Oceania'', is ranked higher in the reasoning process, ensuring that the final CoT provides the most accurate regional classification. This demonstrates that RankCoT is not merely a direct extraction or summary of retrieved documents, but rather a refined reasoning chain that aligns closely with the query.

\end{document}